# A Cost-Effective Thermal Imaging Safety Sensor for Industry 5.0 and Collaborative Robotics

Daniel Barros[1], Paula Fraga-Lamas[2,3], Tiago M. Fernández-Caramés[2,3], and Sérgio Ivan Lopes[1,4,5]

[1] ADiT-Lab, Instituto Politécnico de Viana do Castelo, Viana do Castelo, Portugal
danielbarros@ipvc.pt, sil@estg.ipvc.pt
[2] Department of Computer Engineering, Faculty of Computer Science, Universidade da Coruña, 15071 A Coruña, Spain
paula.fraga@udc.es, tiago.fernandez@udc.es
[3] Centro de Investigación CITIC, Universidade da Coruña, 15071 A Coruña, Spain
[4] CiTin - Centro de Interface Tecnológico Industrial, Arcos de Valdevez, Portugal
[5] IT - Instituto de Telecomunicações, Campus de Santiago, 3810-193 Aveiro, Portugal

**Abstract.** The Industry 5.0 paradigm focuses on industrial operator well-being and sustainable manufacturing practices, where humans play a central role, not only during the repetitive and collaborative tasks of the manufacturing process, but also in the management of the factory floor assets. Human factors, such as ergonomics, safety, and well-being, push the human-centric smart factory to efficiently adopt novel technologies while minimizing environmental and social impact. As operations at the factory floor increasingly rely on collaborative robots (CoBots) and flexible manufacturing systems, there is a growing demand for redundant safety mechanisms (i.e., automatic human detection in the proximity of machinery that is under operation). Fostering enhanced process safety for human proximity detection allows for the protection against possible incidents or accidents with the deployed industrial devices and machinery. This paper introduces the design and implementation of a cost-effective thermal imaging Safety Sensor that can be used in the scope of Industry 5.0 to trigger distinct safe mode states in manufacturing processes that rely on collaborative robotics. The proposed Safety Sensor uses a hybrid detection approach and has been evaluated under controlled environmental conditions. The obtained results show a 97% accuracy at low computational cost when using the developed hybrid method to detect the presence of humans in thermal images.

**Keywords:** Industry 5.0 · Safety · Human-centric · Thermal Imaging · Cobots

## 1 Introduction

With the introduction of new paradigms such as Industry 4.0 [1–4] and Industry 5.0 [5–8] a new focus to improve safety conditions for operators in an industry setting has been made. With Industry 4.0, the focus was on the technological revolution, with the implementation of Cyber-Physical Systems (CPS), Industrial Internet of Things (IIoT) devices, Cloud Computing solutions, or Artificial

Intelligence (AI) enabled systems, with the main goal of making industrial production lines more efficient, flexible and with higher quality standards. Currently, Industry 5.0 emerges as an evolution, unlike the revolutionary Industry 4.0, in the sense that it is focused on human-centricity, sustainability, and resilience [9]. In manufacturing, Industry 5.0 is aimed at combining computation skills with human intelligence and resources in a collaborative setting to further increase a company's value and customer satisfaction [10]. In an environment where humans and collaborative robots must interact and work together, one aspect that cannot be forgotten is safety, both in terms of the safety of the human operators and the safety and integrity of products and production lines.

This paper proposes a cost-effective safety sensor that uses low-cost thermal imaging, to be deployed on factory floors for detecting and counting, in real-time, humans in distinct zones of an image. Thus, enabling safety in Industry 5.0 environments that may include Collaborative Robots (CoBots) and humans in the loop. The adoption of Thermal imaging for this purpose, increases the probability of detection of humans in low visibility conditions, when compared with conventional RGB cameras [11]. Moreover, it increases the system's sensitivity to nonmedical grade human-related activities [12, 13], since it focuses on heat sources detection. However, other redundant safety measures should also be considered. The sensor uses a hybrid method to determine if a human is present in a thermal image and if that human has crossed a virtual fence or is in a specific area. The obtained results show that the developed methods for detecting the human presence in a frame are able to yield a reasonably high accuracy at low computational complexity.

The rest of this paper is structured as follows. Section 2 describes the most relevant works regarding safety sensors for Industry 5.0 and human detection using low-cost thermal imaging. Section 3 presents the proposed system architecture and the devised hybrid detection approach. Section 4 analyzes the obtained results. Finally, Section 5 is dedicated to conclusions and future work.

## 2    Related Works

Multiple works have previously addressed the problem of monitoring and detecting the presence of humans in different environments. For instance, in [14] the authors propose and develop a system capable of detecting humans. For such a purpose they study different identification algorithms (such as HoG, Viola James, and YOLOv3) and analyze the impact of the use of different resolutions on latency time. Specifically, the paper describes a system that consists of four sensors (two eight-megapixel cameras and two ultrasound sensors), which are aimed at being mounted onto the base of a robot. The sensing module of the system uses background subtraction to extract contours and then applies identification algorithms.

Another relevant contribution is presented in [15], where the authors describe a wide-band radar sensor that is continuously monitoring human distance to a collaborative robot. The use of an RF transmitter with large bandwidth results in a high range resolution, allowing the authors to estimate better the distance of a human to the sensor. The presented device, according to the authors, suffers from high complexity, due to the use of high-level signal processing algorithms, reduced spatial resolution and a narrow field of view. Similarly, in [16], the authors present a new method, based on Shape Context Descriptor with Adaboost cascade classifiers, for detecting pedestrians in thermal images. The developed method is proved to have a high human detection rate but suffers from high computational costs.

In [17] it is presented a multi-modal sensor array for safe human-robot interactions. The sensor array consists of ST Micro-electronic VL6180X time-of-flight (ToF) sensors that measure the time the light needs to travel to an object and back to the sensor. For force sensing, eight Melexis MLX90393 sensors are used.

In relation to the used sensors, the authors of [18] present a survey regarding the multiple types of available thermal sensors, including FLIR Lepton 2.0, MLX90640 and others, as well as other sensors commonly used in different industries. Then, the paper discusses the employment of thermal sensors in Heating, Ventilation, and Air Conditioning (HVAC) systems, and for the automotive and manufacturing industries. Also, the authors, during the discussion, present a comparison of different detection algorithms such as AdaBoost, KNN, GNB and SVM, showing that the surveyed sensors are able to estimate occupancy with a 100% accuracy when mounted on the ceiling and a 98.6% accuracy when mounted on a wall. In [12], Braga et al. present the assessment procedure and the results obtained when comparing the performance of two low-cost thermal imaging cameras with a more expensive flagship device for medical screening purposes. Results have shown that, although not being medical grade devices, such cameras can be used for elevated temperature detection events.

Finally, another example of thermal imaging sensors for occupancy estimation is detailed in [22]. The authors compare three thermal sensors (MLX90393, GridEye, and FLIR Lepton 2.0) in terms of their resolution, cost or Field of View (FOV), among other specifications. The presented solution makes use of a combination of active frame analysis, component analysis, feature extraction and classification that can be implemented in various smart building applications.

## 3   Design and Implementation of the Proposed System

The solution presented in this paper is aimed at implementing a cost-effective solution, capable of being installed on factory floors, offices or even outdoors. In addition, the devised system is also intended for being as a one-size-fits-all type of solution, with minimal intervention and adjustment needs, thus being able to operate in different environmental conditions.

The component selection can be divided into three subsystems: processing, sensor interface and thermal image sensor.

- ñ Processing subsystem. For image processing, we opted for Espressif ESP32 [23], due to its low cost, high availability, low power consumption, good efficiency, Bluetooth and Wi-Fi communication capabilities, and also an I2C port for direct interfacing with the sensor interface (i.e., with PureThermal2 module), the power consumption of the ESP32 is about 215 mW in active mode and 19 mW in deep sleep mode [24].
- ñ Sensor interface. For the experiments performed for this paper, the PureThermal2 module was selected due to its low cost and native integration of a Micro-USB port, which enables easy connectivity and is capable of allowing communications via I2C, UART and JTAG [25]. The PureThermal2 module includes a STM32F412 [26], whose firmware may be modified for performing certain processing tasks (e.g., image processing), but, actually, for the work presented in this paper, image processing is performed on the ESP32.
- ñ Thermal image sensor. For the implemented system, the FLIR Lepton 3.5 sensor was selected [27]. Such a sensor is capable of outputting 160x120 pixels images, with thermal sensitivity of less than 50 mK (0.050°C) and a horizontal FOV of 57°. In terms of power consumption, the FLIR Lepton 3.5 consumes nominally 150 mW while operating, 650 mW during the shutter event and 5 mW on standby.

Figure 1 illustrates the conceptual framework, in which it is possible to observe the sensor module used to implement the safety sensor. Additionally, Fig. 2 depicts the operational flowchart of the proposed safety sensor.

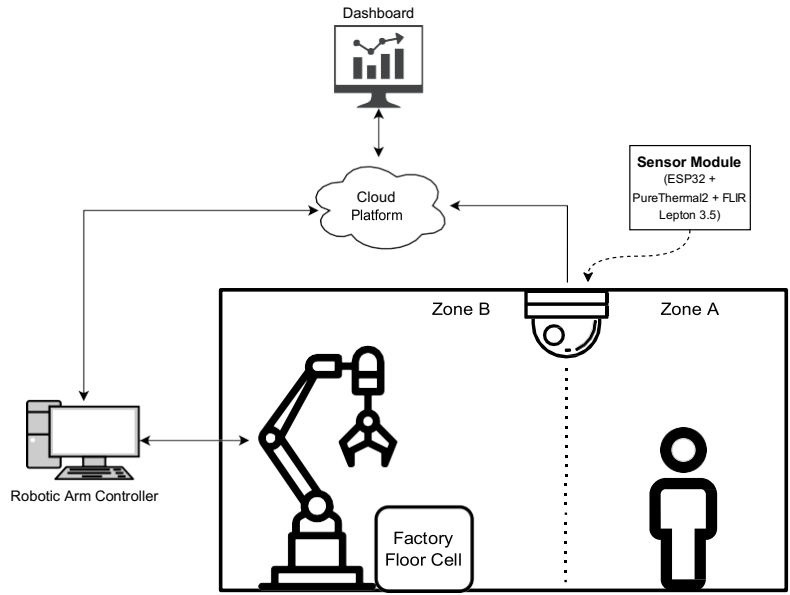

**Fig. 1.** Conceptual framework of the proposed Industry 5.0 use case.

### 3.1   Experimental Testbed

An experimental testbed, cf. Fig. 3, was built to prove whether the use of a thermal imaging sensor would be a capable solution to detect, accurately, persons in a real environment. For such a purpose, a Raspberry Pi 3B was connected via Micro-USB to the PureThermal2 Module with the FLIR Lepton 3.5 camera. Then, the testbed was placed on a vantage point that monitored a research lab, which is frequently used by people. The acquisition of frames was carried out during afternoons, having a large window on the left side of the frame, irradiating heat, which appeared to be negligible due to the high dynamic range of the Flir Lepton 3.5.

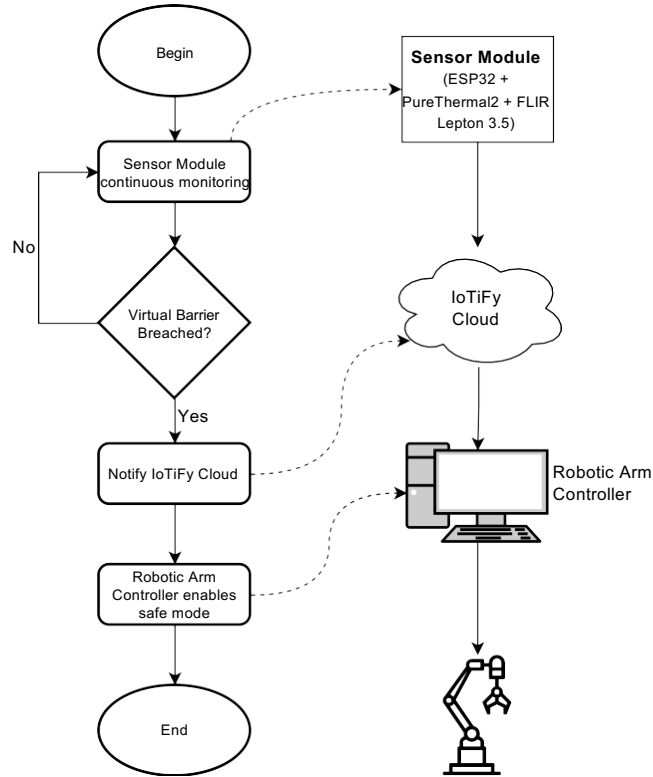

**Fig. 2.** Operational flowchart.

### 3.2 Hybrid Detection Approach

The used detection approach was devised on one simple principle: every human is considered to be a body of heat, and as a such, any body of heat with a shape similar to a human may be considered as one. This principle is only valid with the use of a thermal imaging camera, as a normal RGB camera is not capable of detecting sources of heat.

Considering the mentioned principle, two methods were proposed: one capable of detecting movement and a second that is capable of detecting regions of interest within a frame. Both methods can work simultaneously to determine if there are humans present in an image. If one of the methods determines that there is a human presence, the result is predicted as a positive detection. Only if both methods determine that there is no human present, then the result is negative (i.e., no humans have been detected).

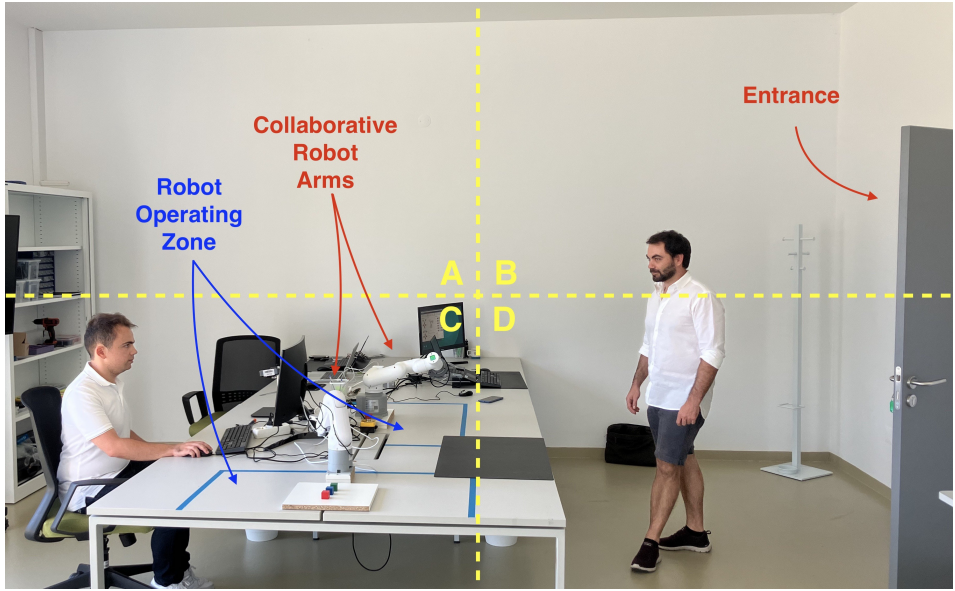

**Fig. 3.** Experimental testbed with users, cobots and the four quadrants identified.

Figure 4 depicts the proposed detection approach consisting of Method A and B. Method A focuses on Movement Detection and consists in the use of two sequential frames: one that will serve as a basis for the comparison and another one that is compared to the previous one.

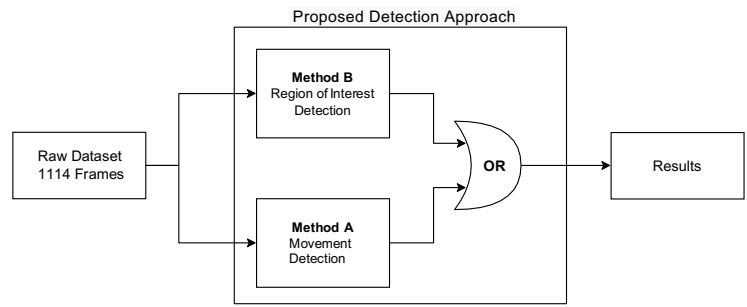

**Fig. 4.** Proposed Detection Approach.

This method uses the direct subtraction of the image matrices to generate a new frame (i.e., the first frame is used as a background, so it is considered as the reference frame). Subsequent frames are then subtracted to the background. This operation has two possible outcomes:

1. The output of the absolute difference results in a new frame with pixel values averaging zero or near zero. This implies that there are no new relevant updates in the new frame, so it can be concluded that there is no movement. In this case, the new image frame will serve as the new background for the comparison with the next frame. This assures that the next comparison is always performed with the most recent reference to compensate for temperature increases during the day and other external factors.
2. If the absolute difference results in a group of pixels (also called active pixels) that have values near a reference value, then a major update is detected, so the likelihood of existing movement is very high. To achieve higher confidence when detecting movement, a comparison needs to be made and it is made using a threshold. Such a threshold is defined as a percentage of active pixels. For the test presented in this paper, the number of pixels in a frame was considered. For example, for a frame of 19,200 pixels (that is the image size for a resolution of 160x120 pixels), if at least 5% of those pixels (i.e., 960 pixels) are considered as active, then it can be determined with certainty that there was movement. If that value is lower than the threshold, then there is no significant movement and no detection is signaled by the algorithm. The percentage of pixels needed for the frame to be considered as a positive is a value that can be adjusted for different conditions or to improve the algorithm sensitivity.

Figure 5 shows the flowchart of the proposed method. Inanimate objects that have heat are discarded with this approach because, although they may have a certain amount of heat that may be identified as a body of heat, when subtracting the sequence of images, only the bodies of heat that moved are shown.

Method B uses a Region of Interest Approach, which consists in using a single 160x120 pixels frame that is then divided into four 80x60 pixels quadrants. Such quadrants do not hold much data on their own, but when compared to the entire frame, they can help to determine if that quadrant has any significance. This comparison is made using the mean value of all the pixels in the total frame with each quadrant, so if the original frame (i.e., entire) has a mean value of $X$, a quadrant will be considered of interest if its mean is at least 20% larger (this value can be adjusted). As it will be later shown in Section 4, this method yields good results for detecting regions of interest in the frame and, when compared with method A, it shows increased accuracy in detecting stationary persons in different parts of the frame. However, one problem that was identified occurs when the body of heat is in the center of the frame, resulting in it being divided into four different quadrants and averaging the values by the frames. Nonetheless, this method can be easily implemented, allowing, if needed, to only monitor a single quadrant (or more) of the image frame. Figure 6 details the flowchart of the proposed method B.

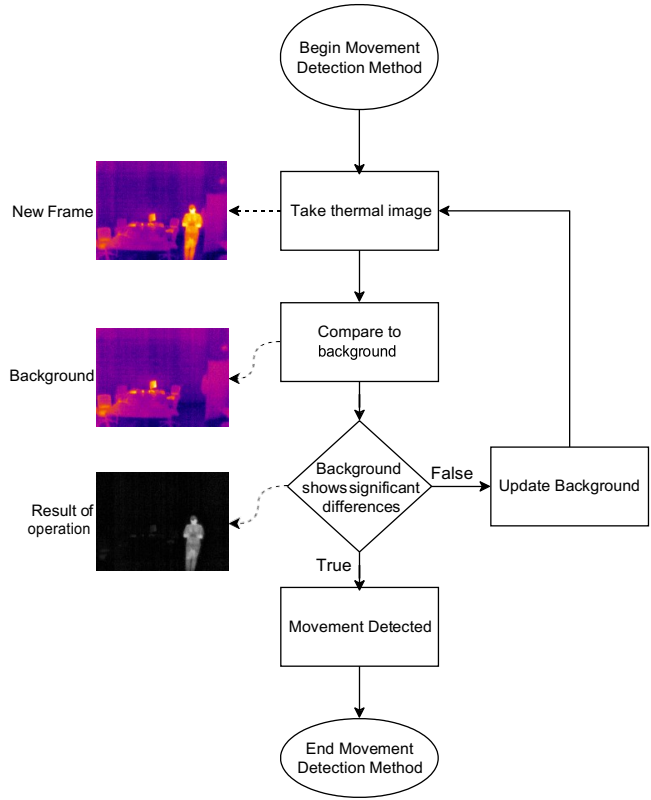

**Fig. 5.** Method A - Movement Detection Approach.

## 4    Results

A script was created for the experimental testbed to first gather the data to be used in the testing of the proposed human-detection methods. Such a script took four consecutive frames every second and was executed continuously for four consecutive days. During that period, 1114 frames were taken and then used to create the reference dataset. Such frames consisted of 1057 frames that contain one or more persons, while the remaining 57 frames had no body of heat present in the frame, except for different types of electrical and electronics equipment, which emitted small portions of heat. Both A and B methods were evaluated individually and combined, as previously described in Fig. 4. The results are presented below in confusion matrixes where the output of each run of the methods is flagged as True Positive (TP), True Negative (TN), False Positive (FP) or False Negative (FN). Then, the accuracy can be obtained using Eq. 1:

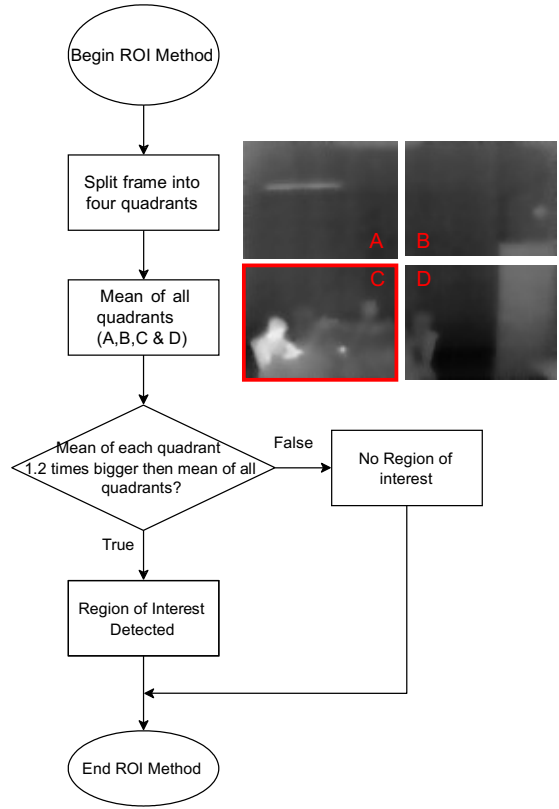

**Fig. 6.** Method B - Region of Interest Approach.

$$((TP + TN)/(TP + TN + FP + FN)) * 100 \qquad (1)$$

Regarding the sole application of method A, its results are shown in the confusion matrix presented in Table 1. The results yielded 1101 positives and 12 negatives, of which 1057 were true positives, 44 were false positives and 12 were false negatives (there were no false negatives). These results show what seems at first a high number of false positives, but this is because of the dataset in which the method is being applied: some frames changed substantially from one day to another, so some new frames were flagged as positives despite being actually negatives. Nonetheless, considering a final Industry 5.0 implementation, having a higher number of false positives can be considered safer than having more false negatives. Overall, method A achieved a 94.5% accuracy with a maximum detection latency of 7 ms per frame.

The results of the application of method B are shown in the confusion matrix presented in Table 2. As it can be observed, 1027 were true positives, 11 were false positives, 48 were true negatives and 28 were false negatives. These results show that method B presents an accuracy using Eq. 1 of 96.5% and a maximum latency of 6 ms per frame. In comparison with method A, method B besides showing a slightly higher percentage of accuracy, displayed more capabilities in detecting stationary people and better performance when dealing with sudden changes in the frame layout.

**Table 1.** Confusion matrix from results of method A.

|  |  |  | Groundtruth (Manually Analized) | |
|---|---|---|---|---|
|  |  |  | 1057 | 57 |
|  |  |  | Positive | Negative |
| Predicted (Method A) | 1101 | Positive | 1057 TP (100 %) | 44 FP (77.2 %) |
|  | 12 | Negative | 12 FN (1.1 %) | 0 TN (0 %) |

**Table 2.** Confusion matrix from results of method B.

|  |  |  | Groundtruth (Manually Analized) | |
|---|---|---|---|---|
|  |  |  | 1057 | 57 |
|  |  |  | Positive | Negative |
| Predicted (Method B) | 1038 | Positive | 1027 TP (97.1%) | 11 FP (19.2%) |
|  | 76 | Negative | 28 FN (2.6%) | 48 TN (84.2%) |

Finally, when considering the hybrid approach, both methods were used simultaneously, first applying method B. Both method A and B used the same parameters as for their individual evaluation. The obtained results indicate that there were 1040 true positive, 41 true negatives, 16 false positives and 17 false negatives, resulting in an overall accuracy of 97.0% and a maximum of 10 ms of detection latency.

It is worth noting that the proposed hybrid approach allows for a good trade-off between performance and computing power, because it does not rely on compute-intense algorithms to identify humans or shapes of humans in a frame, like most Machine Learning (ML) or AI algorithms available. Moreover, both methods A and B only use basic matrix operations to perform the analysis and comparison, and no previous training is required. So 97% accuracy means that this solution can be easily implemented in an edge-like device (e.g., an ESP32) that has low energy consumption, low footprint, low complexity and low cost. Nonetheless, in case of trying to achieve a 99.9% accuracy, some trade-offs would have to be made, mainly in the computational capabilities of the proposed solution.

**Table 3.** Confusion matrix from results of both methods combined.

|  |  |  | Groundtruth (Manually Analized) | |
|---|---|---|---|---|
|  |  |  | 1057 | 57 |
|  |  |  | Positive | Negative |
| Predicted (Both Methods) | 1056 | Positive | 1040 TP (98.4%) | 16 FP (28%) |
|  | 58 | Negative | 17 FN (1.6%) | 41 TN (71.9%) |

## 5  Conclusion and Future Work

This paper described the design and implementation of a real-time cost-effective thermal imaging safety sensor that can be used within the scope of the Industry 5.0 paradigm to trigger different safety states in manufacturing processes that rely on collaborative robotics. Therefore, it enables the protection against possible incidents or accidents or the optimization of the energy efficiency of the deployed industrial devices and machinery. The overall cost of the prototype is approximately e 200, which is roughly the same cost of other solutions that use cloud computing instead of edge computing [19], but more costly when compared to solutions that only use the sensor for gathering data [20, 21], although the latter demand for powerful servers for processing. However, if industrialized, the cost of the proposed solution should reduce considerably, depending on the quantity produced. The proposed method, in contrast to recent literature, avoids relying on complex ML or AI algorithms that require prior training and provides a low energy consumption, low footprint, low complexity and low-cost solution that can be easily deployed instead on a cloud platform on Edge computing devices. The proposed solution can also be deployed with minimal intervention and tuning requirements, thus being able to operate in different environmental conditions. Two approaches were evaluated. First, a movement detection approach achieved a 94.5% of accuracy, and 7 ms maximum detection latency. Second, a region of interest approach achieved a 96.5% of accuracy, and 6 ms of latency. In addition, a hybrid approach that combined both methods simultaneously to determine whether a human is present in a thermal image and whether he/she has crossed a virtual fence or is in a specific area was tested. The results of the hybrid approach showed a promising accuracy of 97%, and 10 ms of maximum latency.

The obtained results were performed with pre-recorded frames. Thus, the next step will consist in using the hardware presented in Section 3 so that the execution of the algorithms can be carried out in real time. From there, with the implemented solution, more analyses and evaluations can be performed, such as the evaluation of the sensor under various failure modes of the components, stress tests, the evaluation of the computational complexity of the pipeline, the assessment of the system latency, to determine the average life time of the sensor, the comparison with other detection algorithms in the same conditions, as well as other solutions using different technologies (e.g., 24 GHz mmWave Sensors [28, 29]). Moreover, determining the long-term maintenance cost of the proposed solution is also interesting as future work.

## Acknowledgment

This work is a result of the project TECH—Technology, Environment, Creativity and Health, Norte-01-0145-FEDER-000043, supported by Norte Portugal Regional Operational Program (NORTE 2020), under the PORTUGAL 2020 Partnership Agreement, through the European Regional Development Fund (ERDF). P.F.L and T.M.F.C. have been supported by Centro de Investigación de Galicia "CITIC" for a three-month research stay in Instituto Politécnico de Viana do Castelo between 15 June and 15 September 2022. CITIC, as Research Center accredited by Galician University System, is funded by "Consellería de Cultura, Educación e Universidades from Xunta de Galicia", supported in an 80% through ERDF Funds, ERDF Operational Programme Galicia 2014-2020, and the remaining 20% by "Secretar´ıa Xeral de Universidades" (Grant ED431G 2019/01). In addition, this work has been funded by the Xunta de Galicia (by grant ED431C 2020/15), the Agencia Estatal de Investigación of Spain, MCIN/AEI/10.13039/501100011033 (by grant PID2020-118857RA-I00 (ORBALLO)) and ERDF funds of the EU (FEDER Galicia 2014–2020 & AEI/FEDER Programs, UE)